# Gaussian Process Topic Models

**Amrudin Agovic**
Dept. of Computer Science & Engineering
University of Minnesota, Twin Cities
Minneapolis, MN 55455
aagovic@cs.umn.edu

**Arindam Banerjee**
Dept. of Computer Science & Engineering
University of Minnesota, Twin Cities
Minneapolis, MN 55455
banerjee@cs.umn.edu

## Abstract

We introduce Gaussian Process Topic Models (GPTMs), a new family of topic models which can leverage a kernel among documents while extracting correlated topics. GPTMs can be considered a systematic generalization of the Correlated Topic Models (CTMs) using ideas from Gaussian Process (GP) based embedding. Since GPTMs work with both a topic covariance matrix and a document kernel matrix, learning GPTMs involves a novel component—solving a suitable Sylvester equation capturing both topic and document dependencies. The efficacy of GPTMs is demonstrated with experiments evaluating the quality of both topic modeling and embedding.

## 1 INTRODUCTION

In recent years, significant progress has been made in analyzing text documents using topic models. Statistical topic models such as Latent Dirichlet Allocation (LDA) [Blei et al.(2003)] and its variants have proven useful and effective. Such topic models allow mixed memberships of documents to several topics, where a topic is represented as a distribution over words.

In LDA, the topic proportions for each document are drawn from a Dirichlet distribution. As a consequence, LDA does not have the flexibility of modeling correlations among the topics. Correlated Topic Models (CTMs) [Blei and Lafferty(2006)] were proposed to address this issue. Instead of a Dirichlet prior, CTMs use a multi-variate normal distribution with a covariance parameter and map samples from the normal distribution to the topic simplex using a mean parameterization. The prior model assumes a fixed mean and covariance parameter for the entire corpus, and the corpus is used to learn these parameters. Correlations between topics are captured by the resulting covariance matrix.

Frequently one might have additional information about a text corpus, possibly in the form of additional features/structures, labels, one or more weighted graphs, etc. For the purposes of this paper, we assume that such additional information can be captured by a suitable kernel defined over the documents. While there has been recent work on incorporating link structure among documents [Chang and Blei(2009)], existing topic models, including CTM, are unable to leverage such information in form of a kernel. In this paper, we propose Gaussian Process Topic Models (GPTMs) which can capture correlations among topics as well as leverage known similarities among documents using a kernel. GPTMs can be considered a generalization of CTMs using ideas from Gaussian Process (GP) embedding and regression. Given a kernel among documents, GPTM defines a Gaussian Process mapping from a suitable document space into the topic space. While topic proportions for all documents in CTM are generated from a single mean, the topic proportions in GPTM are generated from different means. The location of the means for any document is determined by the Gaussian Process mapping.

Gaussian Processes (GPs) define distributions over function spaces and have been successfully used for non-linear regression, classification and embedding [Candela and Rasmussen(2005), Rasmussen and Williams(2005), Lawrence(2003)]. The Gaussian Process Latent Variable Model (GPLVM) [Lawrence(2003)] is a probabilistic embedding method which utilizes a GP mapping from the embedded space to the data space. While GPLVM is powerful non-linear embedding method, current literature does not have effective models for combining kernel based non-linear embedding models such as GPLVM with probabilistic topic models such as CTM. One can obtain embeddings from either family of methods—from LDA/CTM based on the topic structure or from GPLVM using the kernel and observed features. The proposed GPTM can systematically leverage both types of information and obtain an embedding based on both the topic structure and the kernel.

We propose suitable approximate inference algorithms for learning GPTMs and making predictions on a test set. The proposed inference algorithm marginalizes the latent variables in the topic model and maximizes over the latent variables in the embedding, so we obtain one good embedding. During learning, GPTMs work with two different positive definite matrices—a topic covariance matrix over the topics and a document kernel matrix over the documents. Our analysis shows that the two matrices get integrated in an elegant manner to determine the final embedding. In particular, we obtain a Sylvester equation involving both matrices whose solution gives the final embedding. While Sylvester equations have been extensively studied in control theory, to the best of our knowledge, their usage in the context of topic models is novel.

The rest of the paper is organized as follows. We introduce GPTMs in Section 2 and discuss learning GPTMs in Section 3. We present experimental results in Section 4, provide a discussion on our model in Section 5 and conclude in Section 6.

## 2 THE MODEL

Correlated Topic Models (CTMs) [Blei and Lafferty(2006)] are an important recent advance in the realm of topic models [Blei et al.(2003), Griffiths and Steyvers(2004)]. CTMs have the ability to capture correlation among topics. However, CTMs were not designed to capture any additional information regarding the documents, possibly in the form of a kernel over the documents. In this section, we introduce Gaussian Process Topic Models (GPTMs) which are a systematic generalization of CTMs capable of incorporating knowledge from a kernel over the documents.

The key difference between CTM and the proposed GPTM is how the model samples mixed memberships over topics for each document. In CTMs, one samples $\eta \in \mathbb{R}^K$ from a multivariate Gaussian $N(\mu, \Sigma)$ and maps $\eta$ to the topic simplex using a mean parameterization [Blei and Lafferty(2006)]. As a result, $E[\eta] = \mu$, i.e., apriori all documents have the same mixing proportions in expectation. In GPTMs, apriori all documents have different mixing proportions in expectation. The mixing proportions are derived from the kernel over the documents and, intuitively, similar documents according to the kernel have similar mixing proportions.

Given a kernel function $\mathcal{K}$ over documents, the corresponding GP defines a distribution over functions over all documents. For a set of $D$ documents, we get a distribution over $f \in \mathbb{R}^D$ given by

$$p(f|\mathcal{K}) = \frac{1}{(2\pi)^{D/2}|\mathcal{K}|^{1/2}} \exp\left(-\frac{1}{2}f^T\mathcal{K}^{-1}f\right). \quad (1)$$

Assuming there are $K$ topics, we independently sample $f_1, \ldots, f_K \in \mathbb{R}^D$ from the above distribution, and construct a $K \times D$ matrix $F$, whose $i^{th}$ row is $f_i^T$. Hence $p(F|\mathcal{K}) = \prod_{i=1}^{K} p(f_i|\mathcal{K})$. Now, for each document $d = 1, \ldots, D$ we generate $\eta_d \in \mathbb{R}^K$ following

$$p(\eta_d|\mu_d, \Sigma) \sim \mathcal{N}(\eta|\mu_d, \Sigma) \quad (2)$$

where $\Sigma$ denotes a $K \times K$ topic covariance matrix and $\mu_d = Fe_d \in \mathbb{R}^K$, where $e_d \in \mathbb{R}^D$ represents the all zeros vector with only the $d^{th}$ entry 1. Thus, $\mu_d \in \mathbb{R}^K$ is the $d^{th}$ column of $F$. Each $\eta_d$ is then mapped to the topic simplex using the mean parameterization: $\theta(\eta_d) = \frac{\exp(\eta_d)}{\sum_i \exp(\eta_d(i))}$. Since the rows of $F$ were drawn independently from the GP with kernel $\mathcal{K}$, similar documents according to the kernel will implicitly have similar $\mu_d$, and hence similar topic proportions apriori. Thus, the GP prior captures global relationships between documents in determining the apriori mixed memberships.

The entire generative model (Figure 1) can be specified as follows:

1. Draw $F|\mathcal{K} \sim p(F|\mathcal{K}) = \prod_i \mathcal{N}(f_i|0, \mathcal{K})$.

2. For each document $d = 1, \ldots, D$:
   (a) $\eta_d|F, \Sigma \sim \mathcal{N}(\eta_d|Fe_d, \Sigma)$.
   (b) For each word $w_n, n = 1, \ldots, N_d$:
       i. Draw a topic $z_n|\eta_d \sim \text{Discrete}(\theta(\eta_d))$.
       ii. Draw a word $w_n|z_n, \beta_{1:K} \sim \text{Discrete}(\beta_{z_n})$.

The joint probability of all observed and latent variables in GPTM is given by:

$$p(w, z, \eta, F|\mathcal{K}, \Sigma, \beta) = \prod_{i=1}^{K} p(f_i|\mathcal{K}) \prod_{d=1}^{D} p(\eta_d|Fe_d, \Sigma) \prod_{n=1}^{N_d} p(z_n|\eta_d)p(w_n|z_n, \beta). \quad (3)$$

The proposed GPTM is different from both CTM as well as GPLVM, while drawing from the strengths of both of these models. Unlike CTM, the apriori topic proportions of the documents are different and the difference is based on the kernel $\mathcal{K}$. Unlike GPLVM, GPTM takes topic structure into account. The final embedding will be based on both the kernel as well as the structure of the documents as determined by the topic model. As a result, GPTM leverages the strength of both topic models as well as kernel methods, in particular CTMs and GPs.

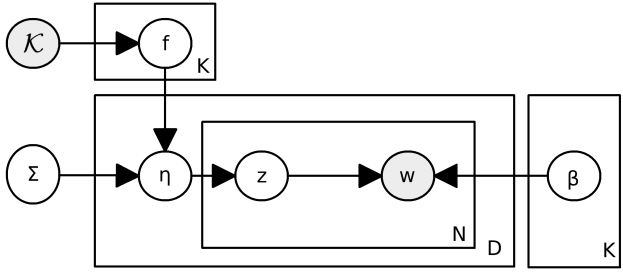

Figure 1: Gaussian Process Topic Model

## 3 LEARNING GPTMs

Exact inference in GPTMs is computationally intractable. In this section, we first explain why some standard approaches to approximate inference may not be desirable and then outline a somewhat nonstandard approach for doing approximate inference in GPTMs. The log-likelihood of the observed words $w$ given $(\mathcal{K}, \Sigma, \beta)$ is given by:

$$\log p(w|\mathcal{K}, \Sigma, \beta) = \log E_{(F,\eta,z)}[p(w, z, \eta, F|\mathcal{K}, \Sigma, \beta)] , \quad (4)$$

where the expectation is over the distribution on the latent variables $(F, \eta, z)$. In several GP-based models [Rasmussen and Williams(2005)], one can integrate over all functions $f$ which translates to the distribution over $F$ in GPTM. Focussing on the terms involving $F$ in the joint distribution, we have

$$\prod_{i=1}^{K} p(f_i|\mathcal{K}) \prod_{d=1}^{D} p(\eta_d|Fe_d, \Sigma) = \frac{1}{|\mathcal{K}|^{k/2}|\Sigma|^{D/2}} \times$$
$$\exp\left\{-\frac{1}{2}\left(\mathrm{Tr}(F\mathcal{K}^{-1}F^T) + \mathrm{Tr}(F^T\Sigma^{-1}F)\right) + \mathrm{Tr}(F^T\Sigma^{-1}\boldsymbol{\eta})\right\},$$

where $\eta = [\eta_1 \cdots \eta_D]$ is the $K \times D$ matrix of means variables $\eta_d$. The terms involving $F$ have both row and column dependencies, one coming from $\mathcal{K}$ and the other from $\Sigma$. Hence, exact marginalization over $F$ is difficult. Further, while the apriori marginal over each entry $\eta_{id}$ of $\eta$ are univariate Gaussians with zero mean and variance $\Sigma_{i,i} + \mathcal{K}_{d,d}$, the joint distribution over $\eta$ will have both row and column dependencies, and hence exact marginalization of $\eta$ is also difficult. While variational inference by assuming a fully factorized distribution over $F, \eta$ will lead to a variational lower bound, such an approach to inference undermines a key property of GPTMs, viz dependencies along both rows and columns. Gibbs sampling based inference is possible for the model but could be computationally burdensome. It involves inverting $(K-1) \times (K-1)$ and $(D-1) \times (D-1)$ matrices for each entry in $\eta$ in each sampling iteration.

In GPTMs, there are two sets of latent variables to consider: the matrix $F$, arising out of the GP, and variables $(\eta, z)$, which are common in topic models [Blei and Lafferty(2006)]. In light of the previous discussion, we choose to maximize the log-likelihood over $F$ and variationally marginalize it over $(\eta, z)$. As we shall see shortly, the maximum aposteriori (MAP) inference over $F$ can be done while maintaining the row and column dependencies. In particular, the first order conditions lead to a Sylvester equation [Wachspress(1988), Bartels and Stewart(1972)] in $F$ involving both $\Sigma$ and $\mathcal{K}$. Further, MAP inference over $F$ leads to an embedding of the data points taking into account both the kernel, the covariance among topics, as well as observed words. While maximizing over $F$ is unconventional in the context of GPs, related ideas have been explored in the recent literature in the context of probabilistic embedding using GPLVMs [Lawrence(2003)]. In Section 4, we compare the embedding performance of GPTMs to that of GPLVMs as well as CTMs.

### 3.1 APPROXIMATE INFERENCE

Our goal in terms of learning is to choose $(F, \Sigma, \beta)$ so as to maximize

$$\log p(w, F|\mathcal{K}, \Sigma, \beta) = \log E_{\eta,z}[p(w, z, \eta, F|\mathcal{K}, \Sigma, \beta)] . \quad (5)$$

In principle, one can also optimize over $\mathcal{K}$ using kernel learning methods, but we do not explore this aspect in this paper. Since computing the expectation over the latent variables $(\eta, z)$ is intractable, following [Blei and Lafferty(2006)], we propose a variational inference approach to lower bound the expectation over $(\eta, z)$. In particular, for each document, we consider the family of fully factored variational distributions $q$ as:

$$q(\eta_{1:K}, z_{1:N}|\lambda_{1:K}, \nu_{1:K}^2, \phi_{1:N}) = \prod_{i=1}^{K} q(\eta_i|\lambda_i, \nu_i^2) \prod_{n=1}^{N} q(z_n|\phi_n) , \quad (6)$$

where $q(\eta_i|\lambda_i, \nu_i^2)$ are univariate Gaussian distributions with mean $\lambda_i$ and variance $\nu_i^2$, and $q(z_n|\phi_n)$ are discrete distributions with parameter $\phi_n$.

Using Jensen's inequality [Blei and Lafferty(2006)], for any $F$ we have:

$$\begin{aligned}
&\log p(w, F|\mathcal{K}, \Sigma, \beta) \\
&\geq \log p(F|\mathcal{K}) + \sum_{d=1}^{D} \Big\{ E_q[\log p(\eta_d|Fe_d, \Sigma)] \\
&\quad + E_q[\log p(z_d|\eta_d)] + E_q[\log p(w_d|z_d, \beta)] \Big\} + H(q)
\end{aligned} \quad (7)$$

We give the exact expressions for each term in Table 1. For the derivation of the last three terms we refer the reader to [Blei and Lafferty(2006)], since these terms are the same as in CTM. The first two terms are unique to our model and stem from the introduction of $F$. For the first term, by definition, we have

$$\log p(F|\mathcal{K})$$
$$= \sum_{i=1}^{K} \left\{ \frac{1}{2} \log |\mathcal{K}^{-1}| - \frac{D}{2} \log 2\pi - \frac{1}{2} f_i^T \mathcal{K}^{-1} f_i \right\}$$
$$= \frac{K}{2} \log |\mathcal{K}^{-1}| - \frac{KD}{2} \log 2\pi - \frac{1}{2} \text{Tr}(F\mathcal{K}^{-1}F^T) \ .$$

For the second term corresponding to each document we have

$$E_q[\log p(\eta_d|Fe_d, \Sigma)] = \frac{1}{2} \log |\Sigma^{-1}| - \frac{K}{2} \log 2\pi$$
$$- \frac{1}{2} E_q[(\eta_d - Fe_d)^T \Sigma^{-1} (\eta_d - Fe_d)]$$
$$= \frac{1}{2} \log |\Sigma^{-1}| - \frac{K}{2} \log 2\pi + \text{Tr}(\text{diag}(\nu_d^2)\Sigma^{-1})$$
$$+ (\lambda_d - Fe_d)^T \Sigma^{-1} (\lambda_d - Fe_d) \ ,$$

where $e_d \in \mathbb{R}^D$ is the all zeros vector with only the $d^{th}$ entry as one. Further, note that the third term $E_q[\log p(z_n|\eta)]$ cannot be computed in closed form, and we obtain a variational lower bound (see Table 1) with parameter $\zeta$ following [Blei and Lafferty(2006)].

### 3.2 PARAMETER UPDATES

The variational lower bound is optimized by updating the variational parameters $(\lambda, \nu, \phi, \zeta)$ and the model parameters $(F, \Sigma, \beta)$. Since parts of our objective function are similar to CTM, a number of updates remain the same [Blei and Lafferty(2006)]. In particular, for the parameters corresponding to the topics, we have:

$$\beta_{i,j} \propto \sum_{d=1}^{D} \sum_{n=1}^{N_d} \phi_{dn,i} w_{dn}^j, \qquad (8)$$

$$\phi_{dj,i} \propto \exp\{\lambda_i\} \beta_{i,j} \ , \qquad (9)$$

where $w_{dn}^j$ is an indicator that the $n^{th}$ word in the $d^{th}$ document is the $j^{th}$ word in the vocabulary. Further, for each document, the update for $\zeta$ is given by:

$$\zeta = \sum_{i=1}^{K} \exp\{\lambda_i + \nu_i^2/2\} \qquad (10)$$

A solution for $\lambda_i$ and $\nu_i^2$ cannot be obtained analytically For each document. So gradient descent is used with gradients

$$g_\lambda = -\Sigma^{-1}(\lambda - Fe_d) + \Sigma_{n=1}^{N} \phi_{n,1:K}$$
$$- (N/\zeta) \exp\{\lambda + \nu^2/2\}$$
$$g_{\nu_i^2} = -\Sigma_{ii}^{-1}/2 - (N/2\zeta) \exp\{\lambda_i + \nu_i^2/2\} + 1/(2\nu_i^2) \ .$$

We now focus on computation of parameters which are different from CTM. Since these are unique to our model, we present them in more detail.

**Computation of $\Sigma$:** Unlike in CTM, we have to compute one covariance matrix given multiple means. Starting with (7) we can pose the problem as:

$$\max_{\Sigma} \left\{ \frac{D}{2} \log |\Sigma^{-1}| - \frac{1}{2} \sum_{d=1}^{D} \text{Tr}(\text{diag}(\nu_d^2)\Sigma^{-1}) \right.$$
$$\left. - \frac{1}{2} \text{Tr}[(L-F)^T \Sigma^{-1} (L-F)] \right\} \ , \qquad (11)$$

with $L = [\lambda_1 \cdots \lambda_D]$ where $\lambda_d$ and $\nu_d$ denote the variational parameters associated with document $d$. Taking the derivative with respect to $\Sigma$, we get

$$\Sigma = \frac{1}{D} \left( \sum_{d=1}^{D} \text{diag}(\nu_d^2) + \sum_{d=1}^{D} (\lambda_d - Fe_d)(\lambda_d - Fe_d)^T \right) \ . \qquad (12)$$

**Computation of F:** The matrix $F \in \mathbb{R}^{K \times D}$ is entirely new in our model. From (7), the optimization problem over $F$ can be posed as:

$$\min_{F} \left\{ \text{Tr}[(L-F)^T \Sigma^{-1}(L-F)] + \text{Tr}[F\mathcal{K}^{-1}F^T] \right\} \qquad (13)$$

Taking derivative with respect to $F$ and setting it to zero, we obtain the following equation:

$$\Sigma F + F\mathcal{K} = \sum_{d=1}^{D} \lambda_d e_d^T \mathcal{K} \ . \qquad (14)$$

With $A = \Sigma$, $B = \mathcal{K}$ and $C = \sum_{d=1}^{D} \lambda_d e_d^T \mathcal{K}$, the equation is of the form: $AF + FB = C$. The equation is known as the Sylvester equation, and it is widely studied in control theory [Golub et al.(1979), Wachspress(1988)]. A solution to the Sylvester equation exists if and only if no eigenvalue of $A$ is equal to the negative of an eigenvalue of $B$. In our case, since $A$ and $B$ are both positive semi-definite, such a situation can arise only if $A$ and $B$ both have at least one zero eigen-value. For that to happen, both $\Sigma$ and $\mathcal{K}$ have to be singular, implying $\Sigma^{-1}$ and $\mathcal{K}^{-1}$ are not well defined. Since $\Sigma$ and $\mathcal{K}$ both act as covariance matrix/function of a Gaussian distribution/process, we assume them to be full rank and positive definite.[1] As a result, a solution to the Sylvester equation exists and can be obtained using standard methods [Wachspress(1988), Bartels and Stewart(1972)].

---

[1]One can generalize the models using pseudo-inserves, but we do not consider such generalizations here.

Table 1: Terms of the lower bound for expected loglikelihood

| Term | Expression |
|------|------------|
| $\log p(F|\mathcal{K})$ | $\frac{K}{2}\log|\mathcal{K}^{-1}| - \frac{KD}{2}\log 2\pi - \frac{1}{2}\text{Tr}(F\mathcal{K}^{-1}F^T)$ |
| $E_q[\log p(\eta_d|Fe_d, \Sigma)]$ | $\frac{1}{2}\log|\Sigma^{-1}| - \frac{K}{2}\log 2\pi - \frac{1}{2}\{\text{Tr}(\text{diag}(\nu^2)\Sigma^{-1}) + (\lambda - Fe_d)^T\Sigma^{-1}(\lambda - Fe_d)\}$ |
| $E_q(\log p(z_n|\eta))$ | $\sum_{i=1}^{K}\lambda_i\phi_{n,i} - \zeta^{-1}\left(\sum_{i=1}^{K}\exp\{\lambda_i + \nu_i^2/2\}\right) + 1 - \log(\zeta)$ |
| $E_q[\log p(w_n|z_n, \beta)]$ | $\sum_{i=1}^{K}\phi_{n,i}\log\beta_{i,w_n}$ |
| $H(q)$ | $\sum_{i=1}^{K}\frac{1}{2}(\log\nu_i^2 + \log 2\pi + 1) - \sum_{n=1}^{N}\sum_{i=1}^{K}\phi_{n,i}\log\phi_{n,i}$ |

### 3.3 INFERENCE ON NEW DOCUMENTS

In the learning phase, one obtains the parameters $(\beta, \Sigma)$ as well as the best $F$ for the training set. While applying the model on new documents, $(\beta, \Sigma)$ will stay unchanged, and we do variational inference to obtain parameters $(\lambda, \nu, \phi, \zeta)$ on the test set. Further, using the fact that location of the mean $\mu_d = Fe_d$ is determined by a GP, we can use GP regression to obtain estimates of document means in the test set.

First, consider one new document, so that the corpus is of size $(D + 1)$. Let $\tilde{F} \in R^{K \times (D+1)}$ denote the matrix containing the means of the entire corpus so that $\tilde{F} = [F \ F_*]$, where $F_*$ denotes the mean for the new document. Let $\tilde{f} = [f \ f_*]$ denote a row of $\tilde{F}$, where $f$ corresponds to the first $D$ documents and $f_*$ corresponds to the new document. A kernel for the entire corpus can be expressed as follows:

$$\tilde{\mathcal{K}} = \begin{bmatrix} \mathcal{K}_{f,f} & \mathcal{K}_{f,*} \\ \mathcal{K}_{*,f} & \mathcal{K}_{*,*} \end{bmatrix},$$

where $\tilde{\mathcal{K}} \in R^{(D+1) \times (D+1)}$. From GP regression [Rasmussen and Williams(2005)], we know that the posterior probability distribution $p(f_*|f)$ can be expressed as:

$$p(f_*|f) = \frac{1}{(2\pi)^{1/2}|\overline{\mathcal{K}}|^{1/2}}\exp\left(-\frac{(f_* - \overline{f})^2}{2\overline{\mathcal{K}}}\right), \quad (15)$$

where $\overline{f} \in \mathbb{R}$ and $\overline{\mathcal{K}} \in \mathbb{R}_+$ is given by

$$\overline{\mathcal{K}} = \mathcal{K}_{*,*} - \mathcal{K}_{*,f}\mathcal{K}_{f,f}^{-1}\mathcal{K}_{f,*} \quad (16)$$
$$\overline{f} = \mathcal{K}_{*,f}\mathcal{K}_{f,f}^{-1}f. \quad (17)$$

Similarly we can obtain a posterior distribution for a collection of $M$ new documents. Let $F_* \in R^{K \times M}$ denote the matrix containing the means of the new documents and $\tilde{F} = [F \ F_*]$. Following the same steps as above, we note that each row $f_{i,*}$ of $F_*$ follows a multi-variate Gaussian distribution with mean equal to the row $\overline{f}_i$ of $\overline{F}$, where $\overline{F}^T = \mathcal{K}_{*,f}\mathcal{K}_{f,f}^{-1}F^T$, and covariance $\overline{\mathcal{K}} = \mathcal{K}_{*,*} - \mathcal{K}_{*,f}\mathcal{K}_{f,f}^{-1}\mathcal{K}_{f,*}$. Since the rows of $F_*$ are independent, the probability of the entire matrix is given by

$$p(F^*|F, \tilde{\mathcal{K}}) = \prod_{i=1}^{K}p(f_{i,*}|F, \tilde{\mathcal{K}}) = \frac{1}{(2\pi)^{K/2}|\overline{\mathcal{K}}|^{K/2}} \times$$
$$\exp\left\{-\frac{1}{2}\text{Tr}\left[(F_* - \overline{F})\overline{\mathcal{K}}^{-1}(F_* - \overline{F})^T\right]\right\}. \quad (18)$$

With the above prior distribution on $F_*$ on the test set conditioned on the $F$ from the training set, the rest of the generative model for the test set remains the same as GPTM. Introducing variational distributions as before, inference on new documents boils down to optimizing the variational parameters and $F_*$ over the test set. The optimization over the variational parameters are same as in the training set. Focussing on the terms involving $F_*$, we get the following problem:

$$\min_{F_*}\left\{\frac{1}{2}\text{Tr}\left[(L - F_*)^T\Sigma^{-1}(L - F_*)\right] \right. \quad (19)$$
$$\left. + \frac{1}{2}\text{Tr}\left[(F_* - \overline{F})\overline{\mathcal{K}}_j^{-1}(F_* - \overline{F})^T\right]\right\},$$

where $L = [\lambda_1 \cdots \lambda_M]$ is the $K \times M$ matrix of variational parameters $\lambda_d$ on the test set. The first order conditions lead to the following matrix equation:

$$\Sigma F_* + F_*\overline{\mathcal{K}} = \left(\sum_{d=1}^{D}\lambda_d e_d^T + \overline{F}\right)\overline{\mathcal{K}}, \quad (20)$$

which is again a Sylvester equation. Note that the right hand side is affected by both observed words in the test set in the form of $\lambda_d$ and by the estimated mean $\overline{F}$ obtained from training set documents.

### 4 EXPERIMENTAL EVALUATION

In this section we evaluate GPTM both as a topic model and as an embedding method, respectively in comparison to CTM and GPLVM.

**Datasets:** Our experiments are performed on 6 text data sets. `Dif100`, `Sim100`, and `Same100` are subsets of the 20Newsgroup data set, each having 300 data points from 3 categories; CMU100 is a larger subset which contains 1000 documents from 10 categories. We also report experiments on NASA's Aviation Safety Reporting System (ASRS) dataset: `ASRS` is a subset of

1000 reports from 25 categories, and `ASRS3` is a subset of 788 documents from 3 categories.

**Kernel Functions:** We consider two kernels for our experiments: an unsupervised nearest-neighbor kernel derived from the document vectors and a semi-supervised must-link kernel derived from must-link constraints on some pairs of documents.

Let $\mathcal{X} = \{x_1, \ldots, x_D\}$ denote a set of feature vectors represented by the word counts in a given document. Let $G = (V, E)$ be a $k$-nearest neighbor graph which is symmetrized by making sure that $(x_i, x_j) \in E$ whenever $(x_j, x_i) \in E$. The graph neighbors are determined using cosine similarity among the document vectors. The nearest neighbor (NN) kernel is defined as:

$$\mathcal{K}_{NN}(x_i, x_j) = \begin{cases} \gamma \exp\left(\frac{-d(x_i, x_j)}{2\sigma^2}\right) & \text{if } (x_i, x_j) \in E \\ c & \text{if } i = j \\ 0 & \text{otherwise}, \end{cases} \quad (21)$$

where $d(x_i, x_j) = 1 - \frac{x_i^T x_j}{\|x_i\|\|x_j\|}$, and $\sigma^2, \gamma, c$ are constant parameters chosen to ensure $\mathcal{K}_{NN}$ is positive definite. The parameters are tuned for each data set using cross validation.

While $\mathcal{K}_{NN}$ can be used in practice, constructing a kernel based on document vectors may not be conceptually desirable from a generative model perspective. Note that the experiments using $\mathcal{K}_{NN}$ demonstrate the ability of our model to incorporate neighborhood information, which may come from supplemental information regarding the documents. Further, GPTMs are not tied to the use of $\mathcal{K}_{NN}$. To illustrate the effectiveness of GPTMs, we consider another kernel constructed purely based on semi-supervised information, viz must-link constraints between certain pairs of documents, without utilizing the documents themselves.

In particular, we consider a semi-supervised setting with a set $\mathcal{C}$ of must link constraints, i.e., if $(x_i, x_j) \in \mathcal{C}$, then the documents are assumed to have the same label. Using such a must-link constraint set, we define the must-link (ML) kernel as:

$$\mathcal{K}_{ML}(x_i, x_j) = \begin{cases} \gamma & \text{if } (x_i, x_j) \in \mathcal{C} \\ c & \text{if } i = j \\ 0 & \text{otherwise}, \end{cases} \quad (22)$$

where $\gamma > 0$ and $c$ is chosen to ensure positive definiteness of $\mathcal{K}_{ML}$. The parameters are tuned per data set using cross validation.

**Perplexity Computation:** In our experiments we use test-set perplexity to evaluate variants of GPTM as well as compare GPTM to CTM. When comparing

Table 2: Perplexity on hold out test set.

|         | CTM          | GPTM $\mathcal{K}_{ML}$ | GPTM $\mathcal{K}_{KNN}$ |
|---------|--------------|-------------------------|--------------------------|
| Dif100  | $1231 \pm 32$ | $1195 \pm 49$          | $\mathbf{1183 \pm 36}$   |
| Sim100  | $1720 \pm 19$ | $1706 \pm 22$          | $\mathbf{1684 \pm 21}$   |
| Same100 | $761 \pm 6$   | $758 \pm 6$            | $\mathbf{755 \pm 12}$    |
| ASRS    | $491 \pm 8$   | $488 \pm 8$            | $\mathbf{483 \pm 3}$     |
| News100 | $2944 \pm 83$ | $2943 \pm 66$          | $\mathbf{2936 \pm 82}$   |

variants of GPTM we compute perplexity as:

$$Perplexity_{w, F_*} = \exp\left(-\frac{\sum_{i=1}^{M} \log p(w_i, F_*)}{\sum_{i=1}^{M} N_i}\right), \quad (23)$$

based on the joint likelihood of the test documents $w_i$ with MAP estimate $F_*$.

To compare our model to CTM, we compute conditional perplexity for $GPTM$ as follows:

$$Perplexity_{w|F_*} = \exp\left(-\frac{\sum_{i=1}^{M} \log p(w_i|F_*)}{\sum_{i=1}^{M} N_i}\right) \quad (24)$$

Since the conditional distribution $p(w|F*)$ is a distribution over the space of documents, which is the same for CTM, the perplexity comparison is meaningful and fair. We also consider additional measures, including topics inferred, document embeddings generated, and classification performance using the embeddings, to get a better understanding of their comparative performance.

### 4.1 GPTM vs. CTM

We report perplexity results comparing CTMs with GPTMs using both the ML-kernel and the NN-kernel with neighborhood $k = 10$ in Table 2. Perplexity was evaluated on a held out test-set using five-fold cross validation and the number of topics set to three. Compared to CTMs, we observe mild to moderate improvements in perplexity across all datasets for GPTMs using both kernels. Thus, in terms of perplexity on the test-set, GPTMs are better or at least as good as CTMs. For GPTMs, the NN-kernel appears to perform mildly better compared to the ML-kernel.

We also qualitatively examined the topics obtained by both models. In Tables 3,4 and 5 we list the 10 most likely words for some of the topics obtained from the `News100` dataset. As is evident, GPTMs returns topics which appear as interpretable as those obtained by CTM. Taking a closer look, we can see that some of the topics appear more coherent in GPTMs. For example, Topic 1 in CTM contains car part related words and the word `convention`. In GPTM with NN-kernel, we have a topic on car parts (Topic 1), and a topic about the liberterian party convention held at the Karl Hess business institute (Topic 2). Topic 2 in CTM appears to be memory related, along with words such as plant, court, cement, and atlantic. While in GPTM with $\mathcal{K}_{NN}$ the corresponding topic (Topic 3) appears only memory related. The two versions of GPTM produce rather similar topics. With $\mathcal{K}_{ML}$ the hockey topic appears more generic, possibly because there are multiple hockey related documents within the same class. While these are only anecdotal examples, the bottom line is that both CTMs and GPTMs produce high quality interpretable topics.

Table 3: Topics extracted by CTM from 20Newsgroup data

| Topic 1 | Topic 2 | Topic 3 | Topic 4 | Topic 5 | Topic 6 |
|---|---|---|---|---|---|
| car | **plant** | echo | pittsburgh | system | god |
| oil | **court** | list | period | mac | existence |
| brake | scsi | motif | lemieux | files | exist |
| fluids | **disk** | xterm | stevens | disk | islam |
| tires | **cement** | set | play | file | standard |
| dot | data | mailing | power | comp | science |
| **convention** | ram | host | njd | software | atheism |
| abs | card | mail | scorer | ftp | religion |
| **braking** | property | sun | pgh | sys | religion |
| cars | **atlantic** | school | islanders | macintosh | laws |

Table 4: Topics extracted by GPTM using $\mathcal{K} = \mathcal{K}_{NN}$ and the 20Newsgroup data

| Topic 1 | Topic 2 | Topic 3 | Topic 4 | Topic 5 | Topic 6 |
|---|---|---|---|---|---|
| car | **convention** | drive | play | system | god |
| oil | don | **disk** | power | mac | good |
| brake | party | hard | period | files | islam |
| fluids | people | drives | pittsburgh | disk | exist |
| tires | price | bios | islanders | comp | pain |
| dot | business | controller | hockey | file | laws |
| abs | hess | floppy | scorer | software | thing |
| cars | karl | card | pts | macintosh | time |
| **braking** | institute | rom | jersey | sys | existence |
| system | libertarian | scsi | good | ftp | faith |

## 4.2 VARIANTS OF GPTMs

We compare four variants of GPTMs to understand the value of the topic covariance matrix $\Sigma$ and the kernel matrix $\mathcal{K}$. In particular, we consider: (i) GPTM-SI-KI, where both $\Sigma$ and $\mathcal{K}$ are identity matrices. Since correlation among topics and similarity among documents are not considered, this model is closest to LDA in spirit; (ii) GPTM-KI, where $\mathcal{K}$ is identity but $\Sigma$ is learnt from the data. This model is closest in spirit to CTM; (iii) GPTM-SI, where $\Sigma$ is identity and $\mathcal{K}$ is set to either $\mathcal{K}_{NN}$ or $\mathcal{K}_{ML}$. This model is similar to LDA with a kernel over documents; and (iv) GPTM, where $\Sigma$ is learned from data and $\mathcal{K}$ is either $\mathcal{K}_{NN}$ or $\mathcal{K}_{ML}$.

Table 2 shows the perplexity numbers on a held out test set using five fold cross-validation. We observe a consistent ordering in terms performance. GPTM-SI-KI performs worst, followed by GPTM-SI, GPTM-KI, and GPTM performs the best. The fact that GPTM-KI performs better than GPTM-SI-KI is consistent with the observation that CTM outperforms LDA. The comparison between GPTM-KI and GPTM-SI shows that GPTM-SI has a consistent better performance possibly implying the kernel adds more value in terms of perplexity than the topic covariance matrix. Finally, the full GPTM outperforms all the special cases. The results clearly illustrate the value in having a suitable kernel over the documents.

Comparing the two different Kernels in GPTM, $\mathcal{K}_{NN}$ fairly consistently results in better perplexity numbers. Interestingly, as far as perplexity is concerned, the nearest neighbor information appears more valuable compared to must-link constraints among the documents.

## 4.3 EMBEDDINGS

We investigate the embeddings obtained by CTM, GPLVM and GPTM using the kNN kernel. With GPLVM a separate derivation of updates is required for different Kernels. We used an existing implementation based on the RBF Kernel. Due to space constraints, we only show results on Dif100 and ASRS3 (Figure 3). The number of topics is set to the correct number of classes for all models and datasets. For CTM we plot $\lambda$, for GPLVM we plot the embedded points and for GPTM we plot $F$. Similar to CTM, for $K$ topics, the degrees of freedom in $\eta$ for GPTMs is $K - 1$ since it eventually gets mapped to the topic simplex. Thus, following CTM, we display the embedding in $(K - 1)$ dimensions. The data points are colored based on their true class label. Note that the CTM embeddings are based on the document structure as provided by the topic model, while GPLVM embeddings are based on the kernel and observed features. GPTM leverages both the topic structure of words as well as the provided kernel.

Table 5: Topics extracted by GPTM using $\mathcal{K} = \mathcal{K}_{ML}$ and the 20Newsgroup data

| Topic 1 | Topic 2 | Topic 3 | Topic 4 | Topic 5 | Topic 6 |
|---|---|---|---|---|---|
| car | liberation | drive | team | system | god |
| brake | committee | tape | hockey | mac | morality |
| oil | institute | scsi | game | files | atheism |
| fluids | president | problem | players | ftp | moral |
| dot | hess | dos | games | file | existence |
| abs | karl | windows | play | disk | exist |
| tires | business | system | year | comp | standard |
| braking | national | floppy | nhl | software | faith |
| system | defense | cable | player | sys | good |
| cars | college | computer | playoffs | macintosh | true |

Table 6: Different Variants of GPTM. Performance improves as more informative kernel and covariance matrices are considered.

| | GP-SIKI | GP-KI | GP-SI $\mathcal{K}_{ML}$ | GP-SI $\mathcal{K}_{KNN}$ | GPTM $\mathcal{K}_{ML}$ | GPTM $\mathcal{K}_{KNN}$ |
|---|---|---|---|---|---|---|
| Dif100 | 1264 ± 47 | 1232 ± 29 | 1198 ± 46 | 1138 ± 42 | 1121 ± 40 | **1095 ± 53** |
| Sim100 | 1778 ± 51 | 1741 ± 51 | 1720 ± 46 | 1656 ± 41 | 1639 ± 46 | **1610 ± 37** |
| Same100 | 829 ± 21 | 792 ± 11 | 745 ± 28 | 633 ± 14 | 638 ± 12 | **608 ± 10** |
| ASRS | 514 ± 11 | 508 ± 7 | 472 ± 7 | 485 ± 9 | **466 ± 9** | 467 ± 7 |
| News100 | 3093 ± 89 | 3047 ± 56 | 2911 ± 65 | 2877 ± 79 | 2818 ± 51 | **2769 ± 86** |

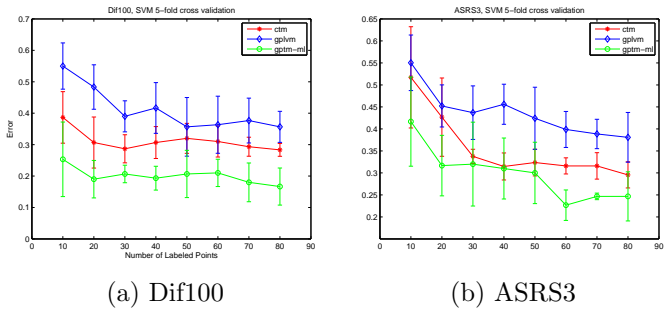

(a) Dif100      (b) ASRS3

Figure 2: The SVM algorithm is applied to the respective outputs of CTM, GPTM and GPLVM.

From Figure 3, we note that for each dataset GPTM produces an embedding where the classes are most cleanly separated. The kernel appears to be helping in preserving the neighborhood structure of the documents which is coherent with the class labels. The better embeddings also help explain why the perplexity goes down when the kNN-Kernel is used.

In Figure 4, we illustrate that the kernel can indeed provide control over document embeddings using the semi-supervised ML kernel defined in (22). We show results on the final embedding using this kernel on Dif100 and ASRS3 using 10, 100, and fully labeled points. Labels are converted to constraints using transitive closure. If we know the class labels (partially) upfront, the question is can we incorporate this knowledge into the topic model? As shown in Figure 4, in each data set, the classes become increasingly separated with additional labeled points while maintaining the structure in each class. The effect is especially evident in Figure 4(f), where the red and blue classes remain somewhat intertwined even with the fully labeled data.

Finally, we apply support vector machines on the embeddings generated by CTM, GPLVM, and GPTM using the must-link kernel. For CTMs and GPLVMs, the partially labeled data is used only for training the SVM. For GPTMs, they are used to determine the ML kernel in GPTM as well as for training the SVM. Due to space constraints, we show results only on Dif100 and ASRS3 in Figure 2. The SVM is trained with an RBF Kernel, and the parameters are tuned using 5-fold cross validation. The classification results show that GPTM produces embeddings with the best class separability and lowest error rates.

## 5 DISCUSSION

In this section we briefly discuss computational aspects in GPTMs and how GPTMs compare to CTMs in terms of capturing topic correlations.

### 5.1 COMPUTATIONAL ASPECTS

There are two major new computational aspects in our model. One pertains to the inversion of the $D \times D$ kernel matrix during GP regression on the test set, and the second is the solution of a Sylvester equation.

The inversion of the kernel matrix is something that most GP based models have to perform [Rasmussen and Williams(2005)]. In recent years, progress has been made on making the computa-tion scalable. For example, several sparsity-based approaches [Lawrence et al.(2002), Foster et al.(2009)] have been developed, which can be readily leveraged in GPTMs. We will explore such approaches in future work.

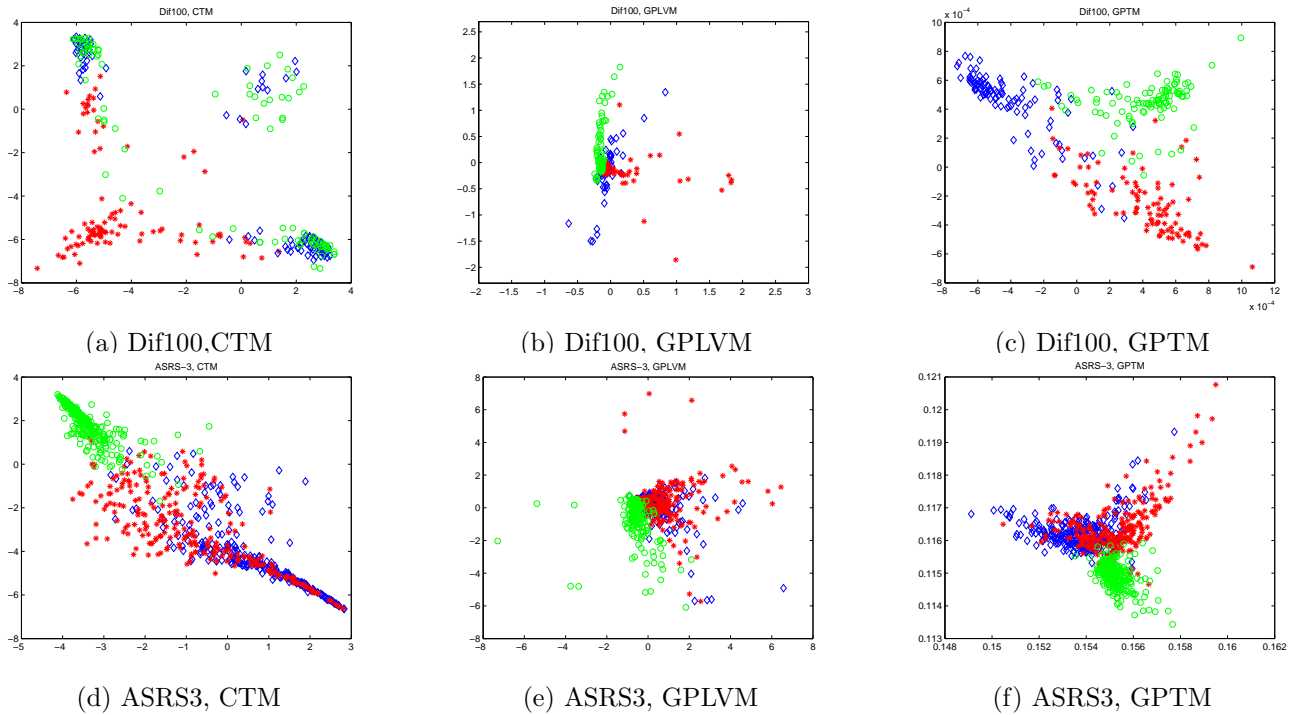

Figure 3: Embeddings obtained from CTM, GPLVM, and GPTM without using class label information. GPTMs separate the classes better than CTM and GPLVM (Best viewed in color).

Solving the Sylvester equation repeatedly is an important computational step in GPTMs. The equation has two matrices: the $K \times K$ topic covariance matrix $\Sigma$ and the $D \times D$ kernel matrix $\mathcal{K}$, where $D \gg K$. During learning, the smaller matrix $\Sigma$ gets updated iteratively whereas the bigger matrix $\mathcal{K}$ is fixed. Using the fact that $\mathcal{K}$ does not change during training, computations involved in solving the Sylvester equation repeatedly can be speeded up significantly. In the test set, since both $\Sigma$ and $\overline{\mathcal{K}}$ are fixed and only the variational parameters $\lambda_d$ on the test set get iteratively updated, repeated solution of the Sylvester equation can also be done efficiently.

### 5.2 TOPIC CORRELATIONS

Throughout this paper we have made comparisons between GPTMs and CTMs. While both models have a topic covariance matrix $\Sigma$, the type of information captured in $\Sigma$ is somewhat different in GPTMs, as a result the covariance matrices in these two models cannot be directly compared.

Consider an example where we model only two topics. A level set of the prior distribution for CTMs is an ellipse over the topic space in $\mathbb{R}^2$. In GPTMs, each document has its own mean in $\mathbb{R}^2$ drawn from the GP. As a result, a level set of the prior for GPTMs involve $D$ ellipses (possibly overlapping/merged) with the same axes but different centroids. In other words, GPTMs consider $D$ different Gaussians with different means but the same covariance matrix. In principle, one can consider different covariance matrices in different parts of the topic space. Such extensions will be considered in future work.

The topic covariance $\Sigma$ in CTMs is inferred only based on the observed words. On the other hand, in GPTMs, the estimated covariance matrix $\Sigma$ also has a dependency on $F$ which in turn depends on the kernel $\mathcal{K}$. While $\Sigma$ in GPTMs may not capture the exact same information as that in CTMs, as shown in Table 6, the performance in terms of perplexity improves when a non-identity covariance matrix is considered.

## 6 CONCLUSIONS

We have introduced a novel family of topic models called GPTMs which can take advantage of both the topic structure of documents and a given kernel among documents. GPTMs can be viewed as a systematic generalization of CTMs which leverages a kernel over documents. The kernel is used to define a GP prior over the topic mixing proportions of documents ensuring that similar documents according to the kernel have similar mixing proportions apriori. The final topic proportions for each document depend both on the kernel as well as the observed words. As our experiments illustrate, with a suitable kernel choice our model can provide good results both in terms of extracted topics as well as the resulting embedding. In particular the kernel allows semi-supervised information to be incorporated into the model, and we illustrate that increased semi-supervision leads to better class separability in the topic space.

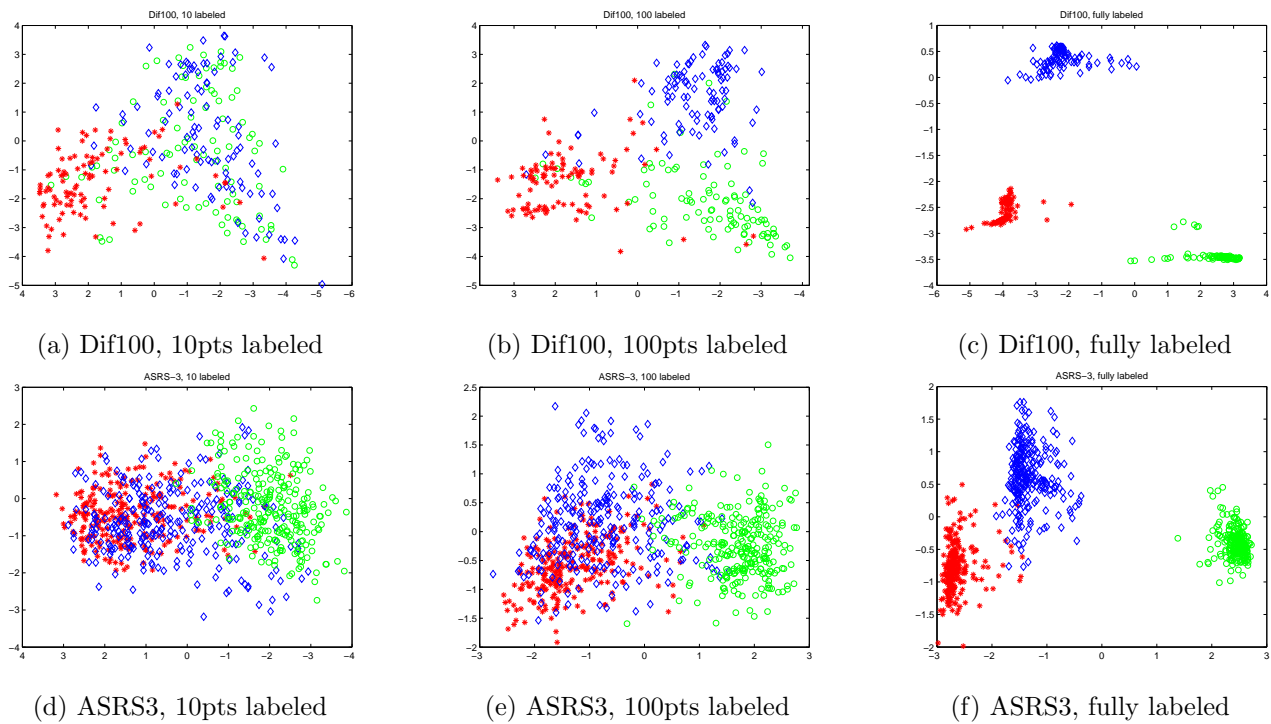

Figure 4: Semi-supervised embeddings from GPTM using ML kernel. As more labeled points are considered, GPTM separates the classes better while preserving topic structure. (Best Viewed in Color)

### Acknowledgements

This research was supported by NSF grants IIS-0916750, IIS-0953274, IIS-0812183, and NASA grant NNX08AC36A.